# Machine-Part cell formation through visual decipherable clustering of Self Organizing Map

Manojit Chattopadhyay[1], Surajit Chattopadhyay[2], Pranab K. Dan[3]
[1,2] Department of Computer Application, Pailan College of Management & Technology, Kolkata-700104
[3] Industrial Engineering, School of Engineering & Technology, West Bengal University of Technology, Kolkata 700 064
[1] chattomanojit@yahoo.com,
[2] surajit_2008@yahoo.co.in
[3] dan1pk@hotmail.com

**ABSTRACT**

Machine-part cell formation is used in cellular manufacturing in order to process a large variety, quality, lower work in process levels, reducing manufacturing lead-time and customer response time while retaining flexibility for new products. This paper presents a new and novel approach for obtaining machine cells and part families. In the cellular manufacturing the fundamental problem is the formation of part families and machine cells. The present paper deals with the Self Organising Map (SOM) method an unsupervised learning algorithm in Artificial Intelligence, and has been used as a visually decipherable clustering tool of machine-part cell formation. The objective of the paper is to cluster the binary machine-part matrix through visually decipherable cluster of SOM color-coding and labelling via the SOM map nodes in such a way that the part families are processed in that machine cells. The Umatrix, component plane, principal component projection, scatter plot and histogram of SOM have been reported in the present work for the successful visualization of the machine-part cell formation. Computational result with the proposed algorithm on a set of group technology problems available in the literature is also presented. The proposed SOM approach produced solutions with a grouping efficacy that is at least as good as any results earlier reported in the literature and improved the grouping efficacy for 70% of the problems and found immensely useful to both industry practitioners and researchers.

## 1. Introduction

Cell formation has been emerged as a production strategy in implementing cellular manufacturing and consists of decomposing the shop in distinct manufacturing cells, each one dedicated to the processing of a family of similar part types. Group Technology (GT) as defined by Burbidge (1979) is the management philosophy that believes similar activities should be done similarly. Cellular Manufacturing (CM) is the application of GT. A manufacturing cell is a cluster of dissimilar machines placed in close proximity and dedicated to the manufacture of a family of parts. In the design of a CM system, similar parts are grouped into families and associated machines into groups as cell formation problem(CF), so that one or more part families can be processed within a single machine group. CM has been proven as a methodology to lower work in process levels, reducing production lead-time while retaining flexibility for new products. The potential benefits include reductions in material handling, setup times, lot sizes, work-in-process inventories and lead times and increase in throughput, productivity and quality (Wemmerlov and Hyer, 1989).

In CF, a binary machine/part matrix of $m \times p$ dimension is usually provided. The m rows indicate m machines and the p columns represent p parts. Each binary element in the $m \times p$ matrix indicates a relationship between parts and machines where ''1'' or ''0'' represents that the pth part should be worked on the mth machine or otherwise. The matrix also displays all similarities in parts and machines. The objective is to group parts and machines in a cell based on their similarities. If there are no ''1'' outside the diagonal block and no ''0'' inside the diagonal block then it is called as perfect result. That is, the two cells are completely independent where each part family will be processed only within a machine group. On the other hand, if in the machine/part matrix there is ''1'' outside the diagonal block then this is called an ''exceptional part'' because it can work on two or more machine groups, and corresponding machine is called a ''bottleneck machine'' as it processes two or more part families. There may also be a ''0'' inside the diagonal block which is called a ''void''. In general, an optimal result for a machine/part matrix by a CF clustering method is desired to satisfy the following two conditions: (a) To minimize the number of 0s inside the diagonal blocks (i.e., voids); (b) To minimize the number of 1s outside the diagonal blocks (i.e., exceptional elements).

There are many cell formation approaches in the literature viz., visual inspection, classification and coding (Singh and Rajamani, 1996), Similarity coefficients(Yin and Yasuda, 2006), Cluster Analysis (Chu and Tsai,1990), array manipulation (Murugan and Selladurai, 2007), Graph Theoretic Approach (Mukhopadhyay, 2009),mathematical programming (Kioon *et al*., 2007), Heuristic algorithms (Mukattash *et al*., 2002), Soft computing technique (Jang *et al*., 2002), Fuzzy clustering (Li, 2007;Tavakkoli-Moghaddam,2007), Metaheuristic techniques like simulated annealing (Lin,2008), genetic algorithm (Pillai *et al*., 2008; Tay and Ho, 2008), Tabu Search (Wang *et al*., 2006), combinatorial search methods(Jeffrey Schaller, 2005), Ants Colony Systems (Kao and Li ,2008), ART1(Yang and Yang,2008;Carpenter and Grossberg, 1987), competitive learning rule (Malave and Ramachandran,1991), Kohonen's self-organizing feature maps (Venkumar P and Haq AN, 2006a,2006b), Fuzzy ART neural network(Suresh *et al*.,1999).

Recently, in most neural network models, competitive learning algorithm (Malave and Ramachandran, 1991) and SOM (chattopadhyay *et al*., 2009; Guerrero *et al*., 2002; Venugopal and Narendran, 1994) has been applied in GT. Since the competitive learning network is an unsupervised approach, it is very suitable for use in GT as the cell formation is a NP complete problem. The robustness of those SOM applications allows it to handle the data visualisation and classification effectively which provides the motivation in applying this approach for exploring into the analysis of the data problem.

In this paper, an attempt is made to use the binary part-machine matrices which are obtained from artificially generated (problem#1 in table 1) and literature (Table 3) to group the parts into part families and machines into machine cells with an idea to maximize the proposed performance measure by introducing a cluster analysis approach using the self organizing map (SOM) proposed by Kohonen (2001). The SOM is a non-linear statistical technique for transforming and visualising multi-dimensional data in a lower-dimensional map (Kohonen, 1998; Mancuso, 2001). The SOM technique based solutions have been designed for problems involving visualisation and cluster analysis (Flexer, 2001; Kiang, 2001) and implemented for many applications for exploratory data analysis. SOM clustering with color coding is a way to group data, according to its properties (Kaski and Kohonen, 1998; Kaski, 2001). The SOM is a competitive learning neural network model, which preserve the distribution and topology information of input data in a low dimension map grid. After mapping, the preserved information can be extracted from which many valuable characteristics of original data can be obtained, such as distribution, cluster, component correlation etc. The novelty of this proposed approach is that the clustering of machine part cell formation can be visually decipherable and thereby cell formation can be studied in depth in a simple and robust way.

All experiments, in the present study were performed in the MATLAB programming language using the SOM MATLAB Toolbox (Vesanto et al., 2000).

## 2. Performance Measure

The grouping efficiency and grouping efficacy are two popular grouping measures because they are simple to implement and generate block diagonal matrices. Grouping efficiency was first proposed by Chandrasekharan and Rajagopolan (1989). It incorporates both machine utilization and inter-cell movement and is defined as the weighted sum of two functions η1 and η2.

The Grouping efficiency (η) is a weighted average of two functions η1 and η2.

$$\eta = (r \times \eta 1) + (1-r)\eta 2 \quad \ldots(5)$$

$$\eta_1 = \frac{\text{Number of ones in the diagonal blocks}}{\text{Total number of elements in the diagonal blocks}} \quad \ldots(6)$$

$$\eta_2 = \frac{\text{Number of zeroes in the off-diagonal blocks}}{\text{Total number of elements in the off-diagonal blocks}} \quad \ldots(7)$$

In equation (5) r is a weighting factor that lies between zero and one (0<r<1) and its value is decided depending on the size of the matrix. A higher value of η is supposed to indicate better clustering.

One drawback of grouping efficiency is the low discriminating capability (i.e. the ability to distinguish good quality grouping from bad). To overcome the low discriminating power of grouping efficiency between well-structured and ill-structured incidence matrices, Kumar and Chandrasekharan (1990) proposed another measure, which they call grouping efficacy. Unlike grouping efficiency, grouping efficacy is not affected by the size of the matrix.

The objective of Eq.(8) is to reach the minimization of the exceptional parts and maximization of the number of parts in cells simultaneously.

The grouping efficacy can be defined as

$$\text{Grouping efficacy} = \mu = \frac{N_1 - N_1^{Out}}{N_1 + N_0^{In}} \quad \ldots(8)$$

Where $N_1$ total number of 1's in matrix; $N_1^{Out}$ total number of 1's outside the diagonal blocks; $N_0^{In}$ total number of 0's inside the diagonal blocks. The closer the grouping efficacy is to 1, the better will be the grouping.

In the present work we have used the grouping efficacy for measuring the performance of cell formation. The grouping efficacy for the matrices obtained from the literature after deploying SOM approaches are compared with the results as reported. The comparisons are given in the table 3 in Appendices.

## 3. Overview of Self Organizing Map Learning Algorithm

Artificial Neural Networks (ANN) is computer algorithm, inspired by the functioning of the nervous system of the human brain, capable of learning from data and generalizing. This learning process can be described as supervised or unsupervised learning. In the supervised learning process, the ANN is shown several input-output patterns during training to enable the trained ANN to make generalizations based on the training data and to correctly produce output patterns based on new input (Jain et al., 1996). The SOM-algorithm is based on unsupervised learning, which means that the desired output is not known a priori. The goal of the learning process is not to make predictions, but to classify data according to their similarity. In the neural network architecture Kohonen proposed (Kohonen, 1998), the classification is done by plotting the data in n-dimensions onto a, usually, two-dimensional grid of units in a topology preserving manner. The former means that similar observations are plotted in each others neighborhood on the 2-D-grid. The neural network consists of an input layer and a layer of neurons. The neurons or units are arranged on a rectangular or hexagonal grid and are fully interconnected. Each of the input vectors is also connected to each of the units. The learning algorithm applied to the network (Kohonen, 2001; Kaski, 1997):

An input vector is shown to the network; the Euclidean distances between the considered input vector $x_i$ and all of the reference vectors $m_i$ are calculated.

$$x_i = [x_1, x_2, ... x_n]^T \in \Re^n, i = 1, 2, ... N \quad ...(1)$$

$$m_i = [m_1, m_2, ... m_n]^T \in \Re^n, i = 1, 2, ... N \quad ...(2)$$

The connection between the two layers represents a map of real high-dimensional data onto a low-dimensional (usually 2-D) display of the nodes. In the training process, The best matching unit m$c$, the unit with the greatest similarity with the considered input vector, is chosen according to:

$$\|x - m_c\| = \min_i (\|x - m_i\|) \quad ...(3)$$

The weights of the best matching unit and the unit within its neighborhood are adapted so that the new reference vectors lie henceforth closer to the input vector. The factor hci(t) controls the rate of change of the reference vectors and is called the learning rate.,

$$m_i(t+1) = m_i(t) + h_{ci}(t)[x(t) - m_i(t)] \quad ...(4)$$

Where t denotes the index of the iteration step, x(t) is the vector-valued input sample of x in the iteration t. Here, the hci(t) is called the neighbourhood function around the winning node c. During training, hci(t) is a decreasing function of the distance between the i-th and c-th model of the map node. For convergence it is necessary that $h_{ci}(t) \to 0$ when $t \to \infty$.

## 4. Visual Decipherable Clustering Approach using Self Organizing Map Method-An Example

A $10 \times 10$ size matrix has been artificially generated by us (problem#1 of table 1) for demonstrating the machine-part cell formation through visual decipherable clustering using our proposed SOM methodology. The SOM methodology was implemented using Matlab programming using somtoolbox and obtained the outputs of umatrix, component planes (figure 1), pc projection (figure 2) and histogram and scatterplots (figure 3). The outputs for the problem#1 of table 1 are discussed below to generate the machine-part cell formation. The block diagonal form (table 2) of the cell formation for the problem#1 was resulted from the output information of the SOM clustering. The details of clustering of machine-part cell formation for machine-part matrix of problem#1 are discussed below.

*4.1 Unified distance matrix (Umatrix)*

The U-matrix makes the 2D visualization of multivariate data possible using codevectors of SOM as source of data. After the learning process is completed, it is derived by using property of topological relations among neurons. This algorithm generates a matrix where each component is a distance measure between two adjacent neurons, therefore we can visualize any multi-variated dataset in a two-dimensional display. Figure 1 shows a representation of a U-matrix calculation for a $12 \times 10$ 2D hexagonal SOM. By U-matrix we can detect topological relations among neurons and infer about the input data structure.

High values in the U-matrix represent a frontier region between clusters, and low values represent a high degree of similarities among neurons on that region, clusters. This can be a visual task when we use some color schema.

*4.2 Component Planes (CP)*

After the learning process we can color each neuron according with each component value in the codevector. For this work we have taken a grey scaled color SOM for each variable (Fig. 1). The emerging patterns of data distribution on SOM's grid (Kohonen, 2001) can be realized through this CP. This can also detect correlations among variables. By only viewing the colored pattern for each CP, the contribution of each one to the SOM differentiation can be well recognised.

*4.3 Visual Inspection of SOM*

The visual inspection of the SOM uses U-matrix on top left, then component planes, and map unit labels on bottom right as shown in figure 1. The maps are connected to adjacent hexagonal nodes with sizes $12 \times 10$, by adapting the observed machines operations('1'). The CP is used for visualising the different input variables representing the operations of machine ('1') identified from the density of colors shades in the nodes network for each CP map. The '1' and '0' correspond to a darker shade, and a lighter shade.

Even a partial correlation of CP may be identified by inspecting grayscales representation from the grid nodes. Each figure in the figure 1, the hexagon in a certain position corresponds to the same map unit. But in the U-matrix, additional hexagons exist between all pairs of neighboring map units. For example, the map unit in top left corner has low values for m1, m3, m5, m9, m10 and relatively high value for m2, m4, m6, m7, m8. The label associated with the map unit is parts and from the U-matrix it can be seen that the unit is very close to its neighbors.

*4.4 Principle Component Projection in SOM Map*

In the figure 2 a principle component projection made to the binary machine-part data set and applied to the map. The SOM grid has been projected to the same subspace. Neighboring map units are connected with lines. The colormap is done by spreading a grey scale colormap on the projection. Distance matrix information is extracted from the U-matrix, and it is modified by knowledge of zero-hits (interpolative) units. Finally, three visualizations are shown: the color code, with clustering information and the number of hits in each unit, the projection and the labels. Blue, green and red, orange correspond to the top and bottom of Umatrix of map respectively and thus extracted two clusters information. Where p1 (blue color) has been extracted with two hits and in reality p2 is similar to p1. The part p3 (red color) is extracted with 6 hits and p4, p6, p7, p8, p9 are similar to p3, p5 (orange color) and p10 (green color) are extracted with a single hit.

*4.5 Scatter Plots and Histograms*

The figure 3 shows the scatter plots and histograms of all variables of machine-part incidence matrix. Original data points are in the upper triangle, map prototype values on the lower triangle, and histograms on the diagonal: black for the data set and red for the map prototype values. The color coding of the parts has been copied from the map (from the BMU of each part). The variable values have been denormalized.

*4.6 Machine-Part Cell formation*

The computational result of the visual decipherable clustering of the machine-part cell formation using our SOM approach for the machine-part incidence matrix problem#1 of table1 are discussed here. The first step in the analysis of the map is visual inspection. Here is the U-matrix, component planes and labels in figure 1.

From this first visualization, we can see that:

- there are essentially two clusters

- m1,m3,m5,m9,m10 are highly correlated

- m2,m4,m6,m7,m8 are also highly correlated

- one cluster corresponds to the p1,p2,p10 and exhibits

- the other cluster corresponds to p3, p4, p5, p6, p7, p8, p9

Next, the projection of the data set is investigated in figure 2. A principle component projection is made for the data, and applied to the map. The colormap is done by spreading a colormap on the projection. Distance matrix information is extracted from the U-matrix, and it is modified by knowledge of zero-hits (interpolative) units. Finally, three visualizations are shown: the color code, with clustering information and the number of hits in each unit, the projection and the labels. From these figures we can see that:

    - the projection confirms the existence of two different clusters

    - interpolative units seem to divide the part family into two classes

Finally, perhaps the most informative figure of all in figure 3: simple scatter plots and histograms of all variables have been derived. The machine cell information is coded as an eleventh variable: m1, m2…m10 for machines. Original data points are in the upper triangle, map prototype values on the lower triangle, and histograms on the diagonal: black for the data set and red for the map prototype values. The color coding of the data samples has been copied from the map (from the BMU of each sample). The variable values have been denormalized. The clear two black blocks inside red map prototype indicates the existence of two clusters of the data set.

This visualization shows quite a lot of information:

distributions of single and pairs of variables both in the data and in the map. If the number of variables was even slightly more, it would require a really big display to be convenient to use.

From this visualization we can conform many of the earlier conclusions, for example:

 - there are two clusters: 'Cell1' (blue, green) and 'Cell2' (orange, reds). This is visible in almost any of the subplots.

 - m1, m3, m5, m9 have a high linear correlation (see subplots 3, 1 and 5, 1 and 9, 1 and also in 5,3 and 9,3 and 9,5)

 - m4, m6, m7, m8 is highly correlated in subplots 6, 4 and 7,4 and 8,4 and 7,6 and 8,6 and 8,7)

 - m2 and m10 have a clear linear correlation, but it is slightly different for the two main clusters (in subplots 10,1 and 10,3 and 10,5 and 10,9 and 4,2 and 6,2 and 7,2 and 8,2).

The part family formation can be similar way analysed and compared the extraction using the color extraction and hit from PC projection of SOM map (figure2) and matched with the U-matrix. Thus two part family with p3, p4, p5, p6, p7, p8, p9 and p1 ,p2, p10 can be formed processed by two machine cells(m1,m3,m5,m9, m10 and m2, m4, m6, m7, m8) as identified by the proposed method of SOM clustering.

These two clusters (Cell1 and Cell2) as derived from all the above discussed visual inspection of SOM map has given rise to two machine-part cells when block diagonal form table 2 has been derived from the original machine-part incidence matrix table 1 after arranging the information of part family processing in machine group.

*4.7 Analysis of Visual decipherable Clustering of cell formation using SOM*

The SOM appears as a robust and flexible clustering method to represent the complexity of the data patterns. For the problem#1 dataset (table 1) the figure 1 shows the U-matrix on top left, then component planes, and map unit labels on bottom right. The twelve figures are linked by position: in each figure, the hexagon in a certain position corresponds to the same map unit. In the U-matrix, additional hexagons exist between all pairs of neighboring map units. For example, the map unit in top left corner has low values for m1, m3, m5, m9, m10 and relatively high value for m2, m4, m6, m7, m8. The label associated with the map unit is parts and from the U-matrix it can be seen that the unit is very close to its neighbors.

The applied dataset (problem#1 in table3) illustrates that SOM method is suitable for extraction of the high dimensional data onto a low dimensional representation. A principle component projection made to the binary machine-part data set and applied to the map. The SOM grid has been projected to the same subspace. Neighboring map units are connected with lines. The colormap is done by spreading a colormap on the projection. Distance matrix information is extracted from the U-matrix, and it is modified by knowledge of zero-hits (interpolative) units. Finally, three visualizations are shown: the color code, with clustering information and the number of hits in each unit, the projection and the labels (figure 2).The SOM map produced an excellent classification and visualisation of machine group via the node clusters. In a different study (Laitinen et al.,2002; Brosse et al.,2001) also reported that SOM proved to be an useful interpretative method for analysis of large size datasets. The most informative aspect of SOM classification is the scatter plots and histograms of all variables in figure 3 for problem#1. Original data points are in the upper triangle, map prototype values on the lower triangle, and histograms on the diagonal: black for the data set and red for the map prototype values. The color coding of the parts has been copied from the map (from the BMU of each part). Note that the variable values have been denormalized. From all of the study above it can be inferred that there are two cells for the problem#1.

Based on the SOM approach of this proposed work the grouping efficacy of problem#1 was 0.97 (problem 1 of table 3). The output as a block diagonal form in cell formation of problem#1 is shown in table 2 which shows that there are no void and 2 exceptional parts and thereby resulted a good cell formation of PMI matrix in table 1.

## 5. Computational Result

To demonstrate the performance of the proposed visual decipherable clustering of Self Organizing Map algorithm, we tested the SOM algorithm on 10 GT instances collected from the literature. The selected matrices range from dimension $5 \times 7$ to $20 \times 35$ and comprise well-structured, as well as unstructured matrices. The matrix sizes and their sources are presented in Table 3. We compare the grouping efficacy obtained by our SOM algorithm with the best grouping efficacies obtained by the seven different approaches (Goncalves and Resende, 2004). These seven approaches provide the best results, found in the literature, for the ten problems used for comparison. The experimental test was run on a personal computer having a Windows XP with Intel® dual-core technology 3.0 GHz processor. The algorithm was coded in Matlab programming. The test results are presented in Table 3. In Appendix B (figure 4) we present the block-diagonal matrices, found after execution of the proposed SOM algorithm, for each of the 10 problems mentioned in Table 3.

As can be seen in Table 3, the SOM algorithm proposed in this paper obtained machine/part groupings, which have a grouping efficacy that is never smaller than any of the best reported results. More specifically, the proposed SOM algorithm obtains for 3 (30%) problems values of the grouping efficacy that are equal to the best ones found in the literature and improves the values of the grouping efficacy for 7 (70%) problems. In 2 (20%) problems, the percentage improvement is higher than 10%.

## 6. Concluding Remarks

A new and novel approach for obtaining machine cells and part families has been presented. The approach proposed a visual decipherable clustering of machine-part cell formation using self organizing map algorithm. Computational experience with the SOM algorithm, on a set of 10 GT problems from the literature, has shown that it performs remarkably well. The algorithm obtained solutions that are at least as good as the ones found the literature. For 70% of the problems, the algorithm improved the previous solutions, in some cases by as much as 11%.The work can be further extended in future incorporating production data like operation sequence, operation time, layout considerations and dynamic manufacturing etc. enhancing it to a more generalized manufacturing environment.

# Appendix A

Table 1 Machine-Part Incidence Matrix problem#1(artificially generated)

|     | m1 | m2 | m3 | m4 | m5 | m6 | m7 | m8 | m9 | m10 |
|-----|----|----|----|----|----|----|----|----|----|-----|
| p1  | 0  | 1  | 0  | 1  | 0  | 1  | 1  | 1  | 0  | 0   |
| p2  | 0  | 1  | 0  | 1  | 0  | 1  | 1  | 1  | 0  | 0   |
| p3  | 1  | 0  | 1  | 0  | 1  | 0  | 0  | 0  | 1  | 1   |
| p4  | 1  | 0  | 1  | 0  | 1  | 0  | 0  | 0  | 1  | 1   |
| p5  | 1  | 1  | 1  | 0  | 1  | 0  | 0  | 0  | 1  | 1   |
| p6  | 1  | 0  | 1  | 0  | 1  | 0  | 0  | 0  | 1  | 1   |
| p7  | 1  | 0  | 1  | 0  | 1  | 0  | 0  | 0  | 1  | 1   |
| p8  | 1  | 0  | 1  | 0  | 1  | 0  | 0  | 0  | 1  | 1   |
| p9  | 1  | 0  | 1  | 0  | 1  | 0  | 0  | 0  | 1  | 1   |
| p10 | 0  | 1  | 0  | 1  | 0  | 1  | 1  | 1  | 0  | 1   |

Table 2 Block Diagonal Form after cell formation of problem#1

|     | m2 | m4 | m6 | m8 | m7 | m1 | m3 | m5 | m9 | m10 |
|-----|----|----|----|----|----|----|----|----|----|-----|
| p2  | 1  | 1  | 1  | 1  | 1  | 0  | 0  | 0  | 0  | 0   |
| p1  | 1  | 1  | 1  | 1  | 1  | 0  | 0  | 0  | 0  | 0   |
| p10 | 1  | 1  | 1  | 1  | 1  | 0  | 0  | 0  | 0  | 1   |
| p3  | 0  | 0  | 0  | 0  | 0  | 1  | 1  | 1  | 1  | 1   |
| p4  | 0  | 0  | 0  | 0  | 0  | 1  | 1  | 1  | 1  | 1   |
| p8  | 0  | 0  | 0  | 0  | 0  | 1  | 1  | 1  | 1  | 1   |
| p9  | 0  | 0  | 0  | 0  | 0  | 1  | 1  | 1  | 1  | 1   |
| p6  | 0  | 0  | 0  | 0  | 0  | 1  | 1  | 1  | 1  | 1   |
| p7  | 0  | 0  | 0  | 0  | 0  | 1  | 1  | 1  | 1  | 1   |
| p5  | 1  | 0  | 0  | 0  | 0  | 1  | 1  | 1  | 1  | 1   |

Table 3. Computational results

| Problem | Source | Size | Best Result of Group efficacy from Other Approaches | Result of Group efficacy using Our Approach | % Improvement in Group efficacy |
|---------|--------|------|------|------|------|
| 1  | Problem#1(table 1) | $10 \times 10$ | --- | 0.97 | --- |
| 2  | Mosier and Taube(1985) | $10 \times 10$ | 70.59 | 70.59 | 0.00 |
| 3  | Carrie(1973) | $20 \times 35$ | 76.22 | 77.91 | 2.22 |
| 4  | Waghodekar and Sahu(1984) | $5 \times 7$ | 62.50 | 69.57 | 11.31 |
| 5  | Balasubramanian and Panneerselvam(1993) | $15 \times 10$ | 72.10 | 75.00 | 4.02 |
| 6  | Balakrishnan and Jog(1995) | $12 \times 19$ | 56.00 | 58.59 | 4.63 |
| 7  | Chandrasekharan and Rajagopalan(1986) | $8 \times 20$ | 85.25 | 85.25 | 0.00 |
| 8  | Seiffodini (1989) | $5 \times 18$ | 68.3 | 77.36 | 13.27 |
| 9  | Chan *et al.* (1982) | $10 \times 15$ | 92 | 92 | 0.00 |
| 10 | Askin *et al.* (1987) | $14 \times 24$ | 64.36 | 66.67 | 3.59 |
| 11 | Stanfel (1985) | $14 \times 24$ | 67.11 | 69.33 | 3.31 |

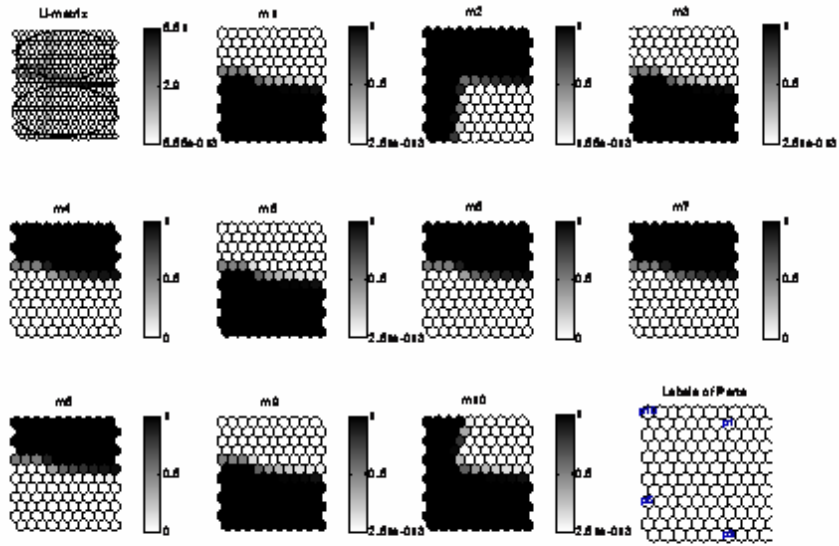

Figure 1. Visualization of the SOM of binary machine-part data.

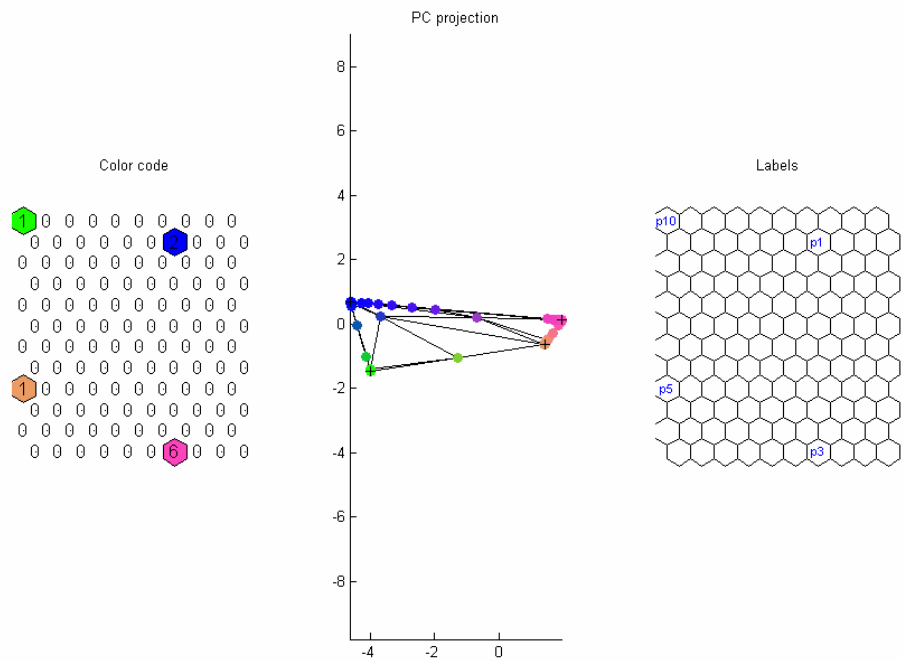

Figure 2. A principle component projection made to the binary machine-part data set and applied to the map.

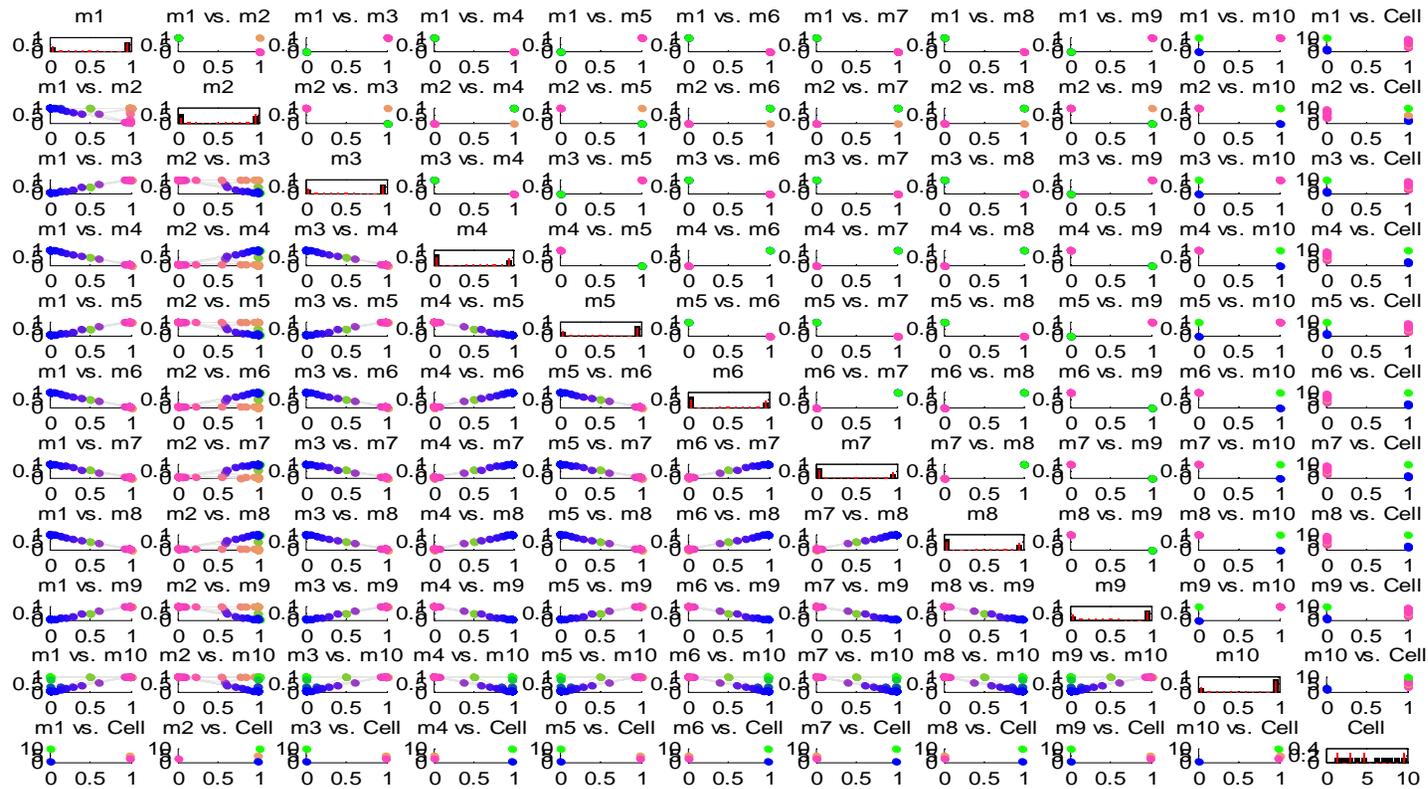

Figure 3. Scatter plots and histograms of all variables. Original data points are in the upper triangle, map prototype values on the lower triangle, and histograms on the diagonal: black for the data set and red for the map prototype values. The color coding of the parts has been copied from the map (from the BMU of each part). Note that the variable values have been denormalized.

# Appendix B

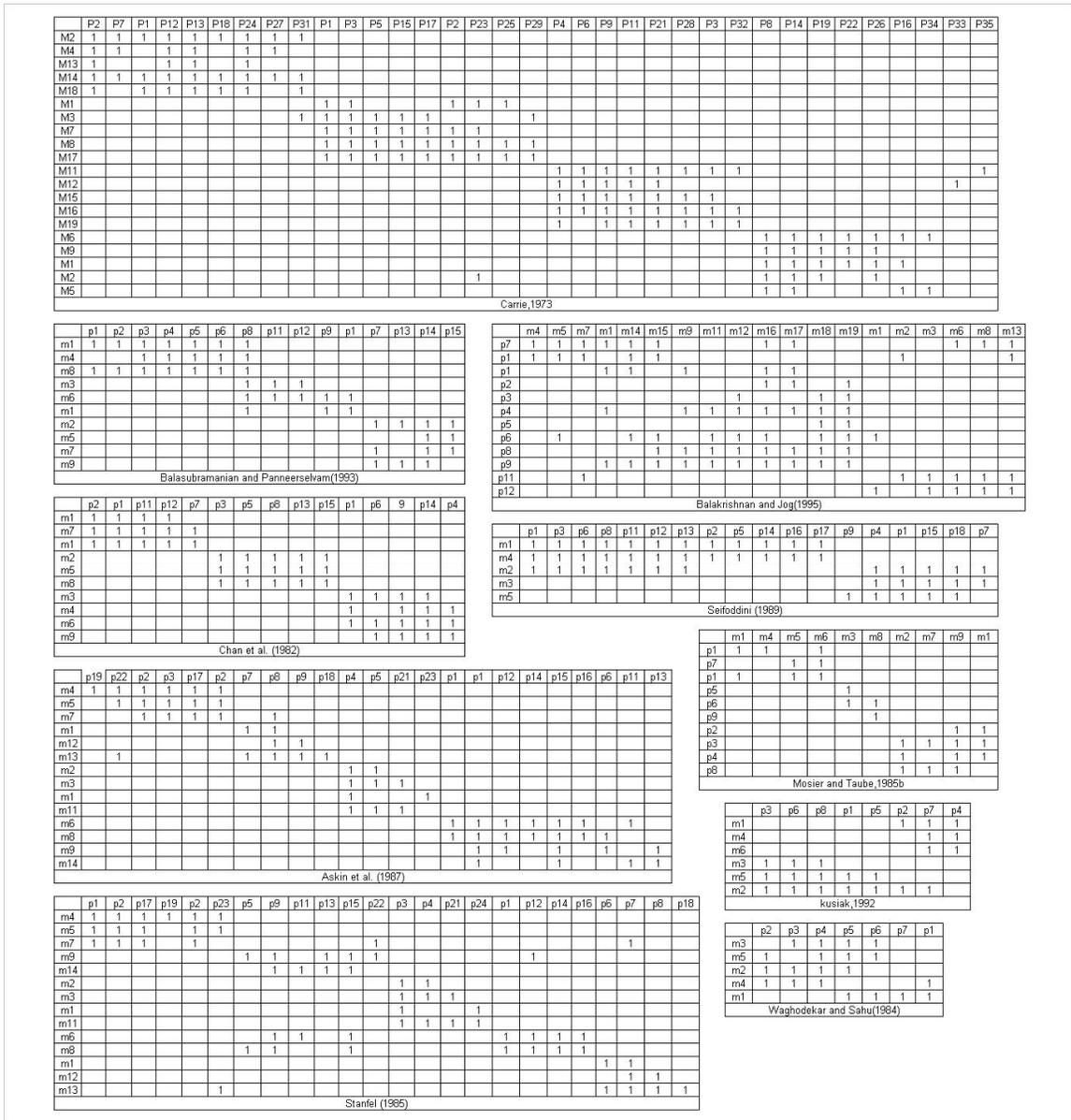

Figure 4. Block Diagonal Result of the machine-part incidence matrices after implementing the Proposed SOM algorithm